\documentclass{article}
\usepackage{microtype}
\usepackage{graphicx}
\usepackage{subcaption}
\usepackage{booktabs} 
\PassOptionsToPackage{round}{natbib}

\usepackage{hyperref}

\newcommand{\improve}[1]{\textcolor{green!70!black}{(#1)}}

\usepackage[preprint]{icml2026}

\usepackage{amsmath}
\usepackage{amssymb}
\usepackage{mathtools}
\usepackage{amsthm}
\usepackage{enumitem}

\usepackage[capitalize,noabbrev]{cleveref}

\theoremstyle{plain}

\theoremstyle{definition}

\theoremstyle{remark}

\usepackage{multirow}
\usepackage[table]{xcolor}
\newcolumntype{O}{>{\columncolor[HTML]{DCEEFF}}c}
\usepackage[textsize=tiny]{todonotes}

\icmltitlerunning{Why Diffusion Language Models Struggle with Truly Parallel
(Non-Autoregressive) Decoding?}

\begin{document}

\twocolumn[
  \icmltitle{Why Diffusion Language Models Struggle with Truly Parallel (Non-Autoregressive) Decoding?}

  \icmlsetsymbol{equal}{*}

  \begin{icmlauthorlist}
    \icmlauthor{Pengxiang Li}{hkpoly,equal}
    \icmlauthor{Dilxat Muhtar}{ellis,mpi,tai,equal}
    \icmlauthor{Tianlong Chen}{unc}
    \icmlauthor{Lu Yin$^\dagger$}{surrey}
    \icmlauthor{Shiwei Liu$^\dagger$}{ellis,mpi,tai}
  \end{icmlauthorlist}

  \icmlaffiliation{hkpoly}{The Hong Kong Polytechnic University, Hong Kong, China}
  \icmlaffiliation{ellis}{ELLIS Institute Tübingen, Tübingen, Germany}
  \icmlaffiliation{mpi}{Max Planck Institute for Intelligent Systems, Tübingen, Germany}
  \icmlaffiliation{tai}{Tübingen AI Center, Tübingen, Germany}
  \icmlaffiliation{surrey}{University of Surrey, Guildford, United Kingdom}
  \icmlaffiliation{unc}{
The University of North Carolina at Chapel Hill, 
Chapel Hill, NC, USA
}
    \icmlcorrespondingauthor{Lu Yin}{l.yin@surrey.ac.uk}

  \icmlcorrespondingauthor{Shiwei Liu}{sliu@tue.ellis.eu}

  \icmlkeywords{Machine Learning, ICML}

  \vskip 0.3in
]



\printAffiliationsAndNotice{}  

\begin{abstract}
Diffusion Language Models (DLMs) are often advertised as enabling parallel token generation, yet practical ``fast'' DLMs frequently converge to left-to-right, autoregressive (AR)-like decoding dynamics. In contrast, genuinely non-AR generation is promising because it removes AR’s sequential bottleneck, better exploiting parallel hardware to reduce synchronization/communication overhead and improve latency scaling with output length. We argue that a primary driver of AR-like decoding is a mismatch between DLM objectives and the highly sequential structure of widely used training data, including standard pre-training corpora and long chain-of-thought (CoT) supervision. Motivated by this diagnosis, we propose \textbf{NAP} (Non-Autoregressive Parallel DLMs), a proof-of-concept, data-centric approach that better aligns supervision with non-AR parallel decoding. NAP curates examples as multiple independent reasoning trajectories and couples them with a parallel-forced decoding strategy that encourages multi-token parallel updates. Across math reasoning benchmarks, NAP yields stronger performance under parallel decoding than DLMs trained on standard long CoT data, with gains growing as parallelism increases. Our results suggest that revisiting data and supervision is a principled direction for mitigating AR-like behavior and moving toward genuinely non-autoregressive parallel generation in DLMs. Our code is available at \href{https://github.com/pixeli99/NAP}{https://github.com/pixeli99/NAP}.

\end{abstract}

\vspace{-2em}
\section{Introduction}

Large language models (LLMs) have become a cornerstone of modern AI, yet their rapidly growing computational and environmental footprints raise pressing sustainability concerns \citep{patterson2021carbon,luccioni2023estimating}. This motivates renewed interest in alternative generation paradigms that can reduce inference latency and cost without sacrificing capability. \emph{Diffusion Language Models} (DLMs) have recently emerged as a compelling candidate: by iteratively denoising a sequence, DLMs can in principle enable \emph{parallel token generation}, offering a path toward faster, more efficient generation \cite{austin2021structured,lou2023discrete,shi2024simplified,sahoo2024simple,nie2025large,gong2024scaling,ye2025dream}. 
When paired with established inference accelerators, such as \emph{KV caching} \citep{ma2025dkvcachecachediffusionlanguage,wu2025fast,liu2025dllm} and \emph{speculative decoding} \citep{christopher2025speculative,gao2025self,chen2026dflash}, DLM-based systems are often claimed as substantially faster alternatives to standard autoregressive (AR) decoding.

\begin{figure*}[h]
    \centering
    \begin{subfigure}[t]{0.24\textwidth}
        \centering
        \includegraphics[width=\textwidth]{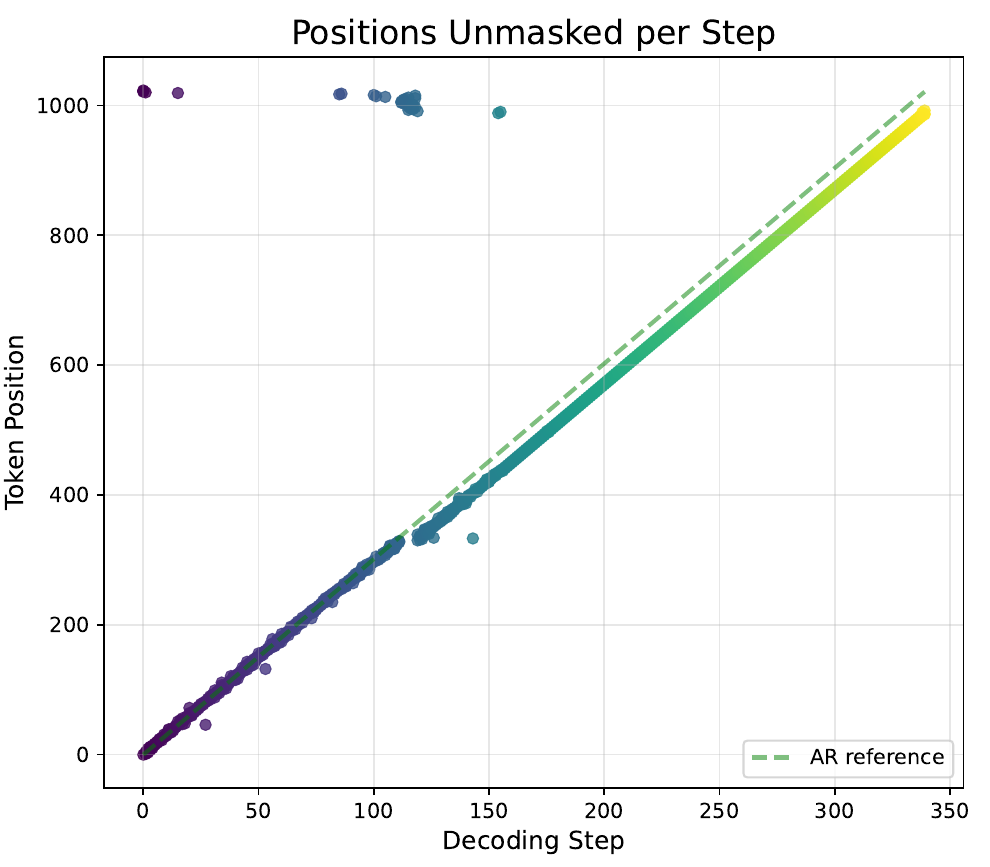}
        \caption{LLaDA-8B (AO)}
        \label{fig:llada}
    \end{subfigure}
    \hfill
    \begin{subfigure}[t]{0.24\textwidth}
        \centering
        \includegraphics[width=\textwidth]{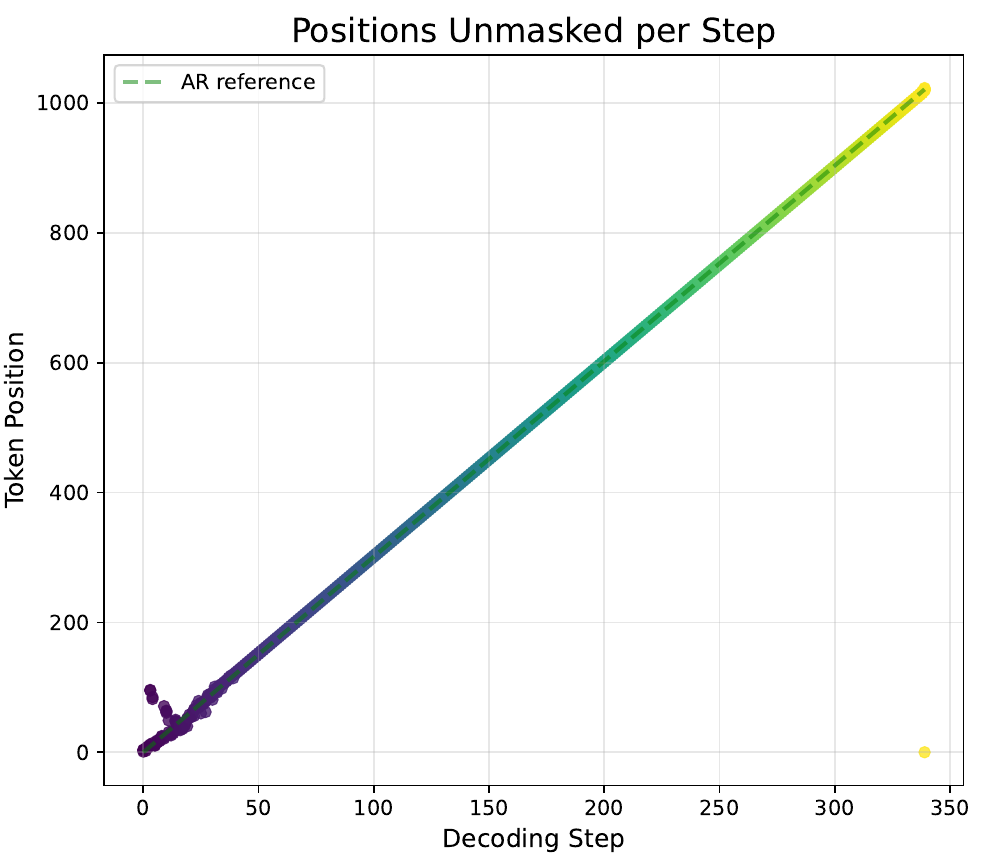}
        \caption{Dream-7B (AO)}
        \label{fig:dream_ao}
    \end{subfigure}
    \hfill
    \begin{subfigure}[t]{0.24\textwidth}
        \centering
        \includegraphics[width=\textwidth]{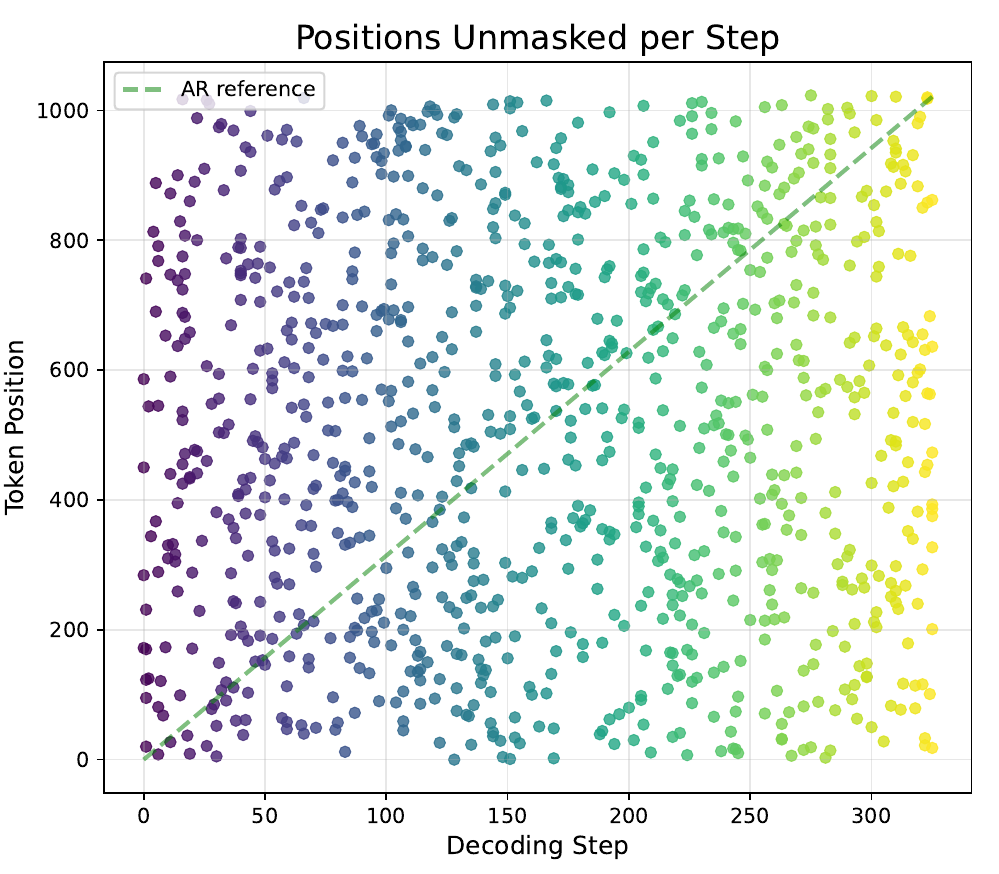}
        \caption{Random}
        \label{fig:dream_rand}
    \end{subfigure}
    \hfill
    \begin{subfigure}[t]{0.24\textwidth}
        \centering
        \includegraphics[width=\textwidth]{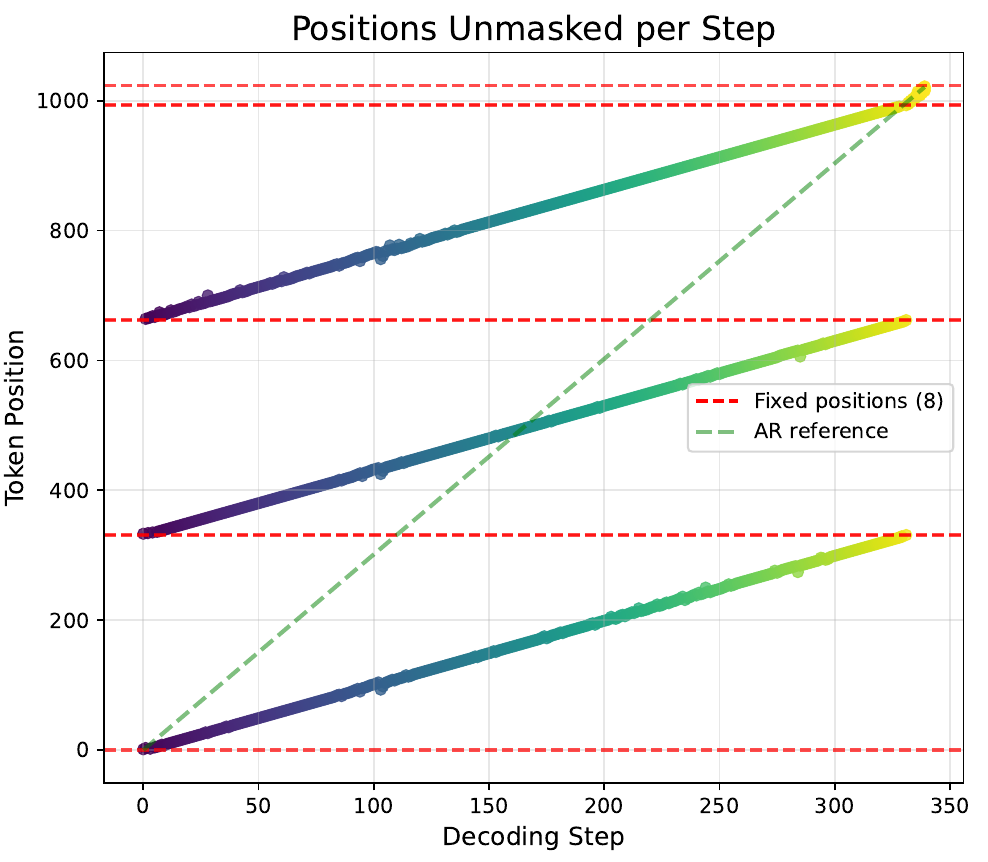}
        \caption{Ours}
        \label{fig:ours}
    \end{subfigure}

    \caption{
Visualization of decoding dynamics. We plot the token position being unmasked (y-axis) against the decoding step (x-axis).
\textbf{(a, b)} Despite using confidence-based Arbitrary Order (AO) decoding, standard DLMs (LLaDA and Dream) exhibit a strict linear diagonal pattern, revealing that their behavior collapses into autoregressive (left-to-right) generation.
\textbf{(c)} Random decoding eliminates AR bias but lacks structure.
\textbf{(d)} Our method (NAP) breaks the single-stream bottleneck, generating multiple reasoning trajectories simultaneously.
}
    \label{fig:decoding_comparison}
\end{figure*}
Yet, despite their promise, practical ``fast'' DLMs exhibit a striking and under-discussed behavior: many methods that aim for highly parallel decoding \emph{converge toward AR-like generation}, where the effective reasoning trajectory proceeds largely \emph{from left to right} \citep{nie2025large,DBLP:journals/corr/abs-2506-00413,wu2025fast,gong2025diffucoder}. In other words, even when the model architecture permits bidirectional context and parallel refinement, the realized decoding dynamics can resemble a sequential construction of the output. This phenomenon makes real-world DLM usage more nuanced than the headline promise of ``truly parallel decoding'': speedups are often coupled to subtle quality trade-offs, and the conditions under which DLMs depart meaningfully from AR behavior remain unclear \cite{kang2025parallelbench}. 

\textbf{\emph{The payoff for achieving genuinely (non-AR) parallel decoding is substantial}}: AR-style decoding is fundamentally sequential, every token depends on the previous one, so generation latency scales roughly with output length. Although we can switch to fast parallel decoding in subsequent blocks after earlier blocks have largely converged, the need to wait for upstream stabilization introduces a sequential critical path, leading to extra latency and communication cost \cite{wang2025diffusion,fu2025bits}. In contrast, truly non-AR parallel decoding is naturally compatible with the distributed hardware, i.e., when dependencies across spans are weak, decoding is naturally compatible with distributed hardware and can be distributed across devices, with only occasional synchronization to maintain global consistency.

In this work, we argue that one primary caveat of this AR-bias is a \textbf{\emph{mismatch between the learning objective and the training data}}. Existing DLM pipelines blindly reuse training data originally designed for AR models, where reasoning trajectories are implicitly encoded as left-to-right progressions, e.g., next-token prediction--style ordering \cite{ye2025dream,allal2025smollm2,li2024datacomp}, or sequential Chain-of-Thought (CoT) rationales \citep{DBLP:journals/corr/abs-2504-12216,lambert2024tulu}. As a result, even if the diffusion process is nominally position-agnostic, the model can learn denoising strategies that preferentially reconstruct outputs in an AR-shaped manner. This ``AR-shaped data'' effect not only limits the extent to which DLMs can exploit genuine parallelism, but also complicates evaluation: a method may appear effective while largely reproducing AR model's dynamics under a different wrapper.

\begin{figure}[t]
    \centering
    \includegraphics[width=\linewidth]{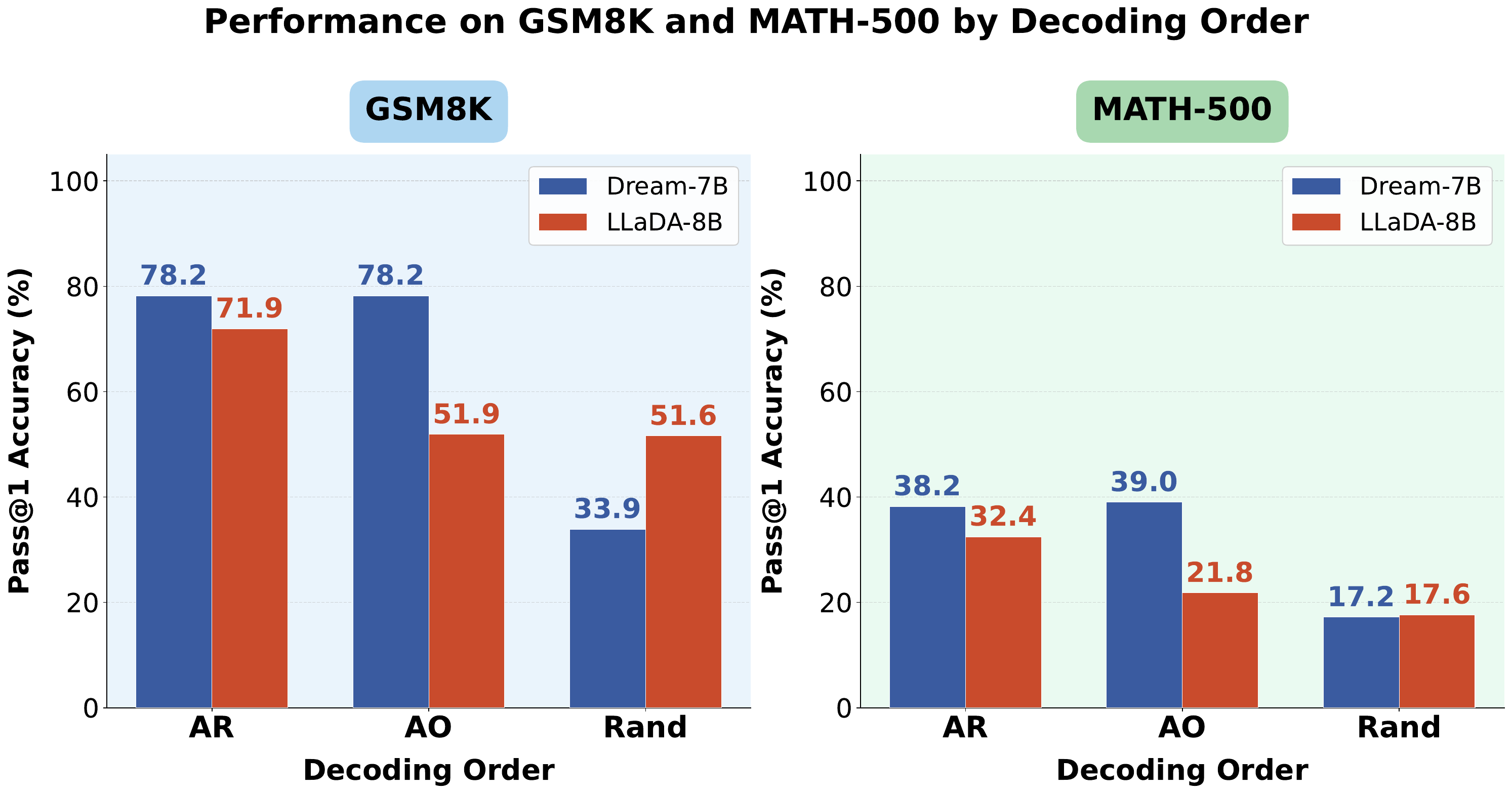}
    \caption{
    \textbf{Performance on GSM8K (left) and MATH-500 (right).}
    Forcing low-ARness behavior (Random decoding) generally causes reasoning performance to collapse. Notably, for LLaDA, we employ a constrained block-wise decoding strategy to ensure generation validity. This preserves local structural integrity, resulting in the Arbitrary Order (AO) decoding maintaining comparable performance, unlike the sharp drop observed in fully unstructured random decoding. 
}
    \label{fig:teaser_accuracy}
\end{figure}

\begin{figure*}[t]
    \centering
    \begin{subfigure}[t]{0.49\linewidth}
        \centering
        \includegraphics[width=\linewidth]{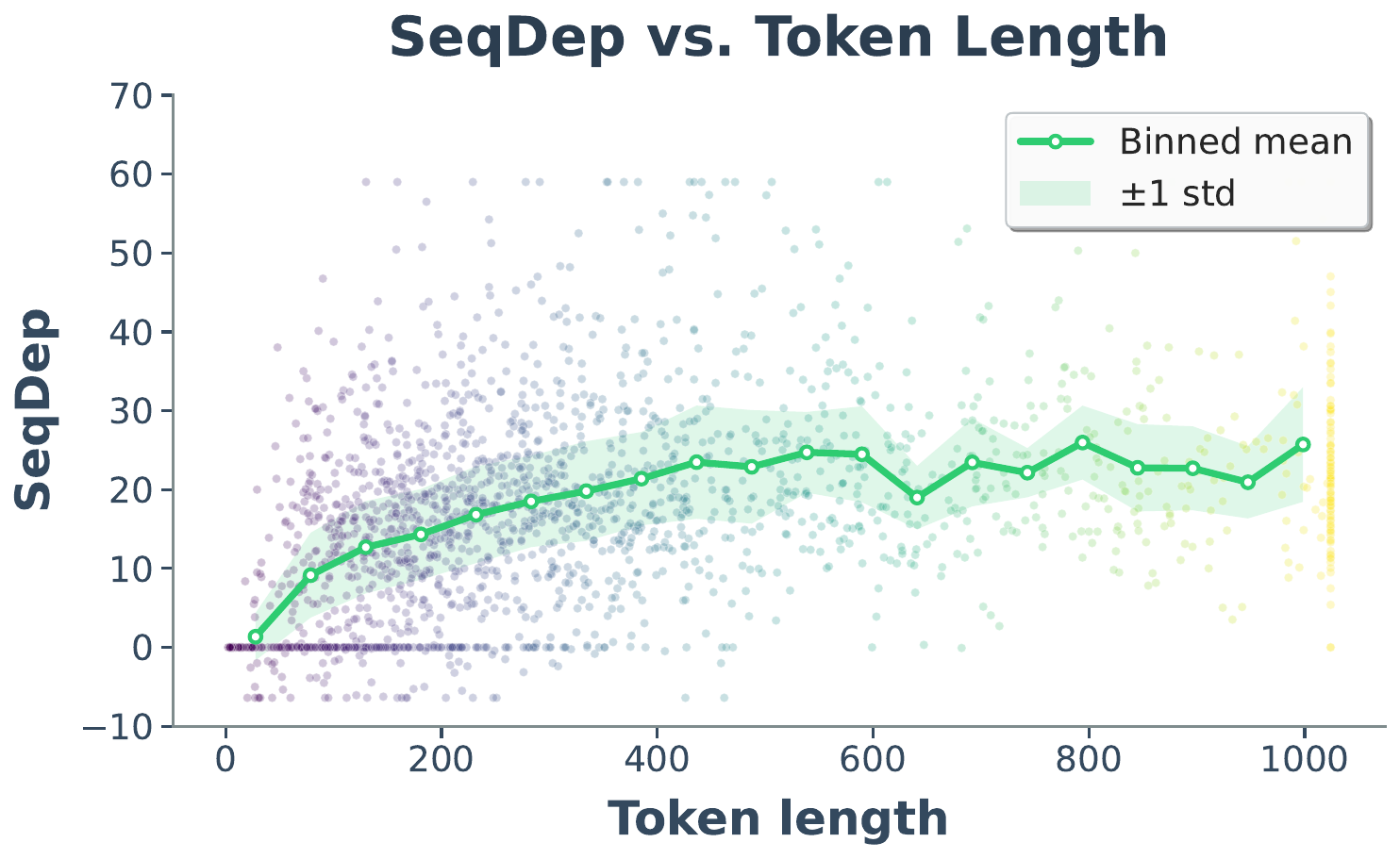}
        \caption{OpenR1-Math SeqDep}
        \label{fig:seqdep_openmath}
    \end{subfigure}
    \hfill %
    \begin{subfigure}[t]{0.49\linewidth}
        \centering
        \includegraphics[width=\linewidth]{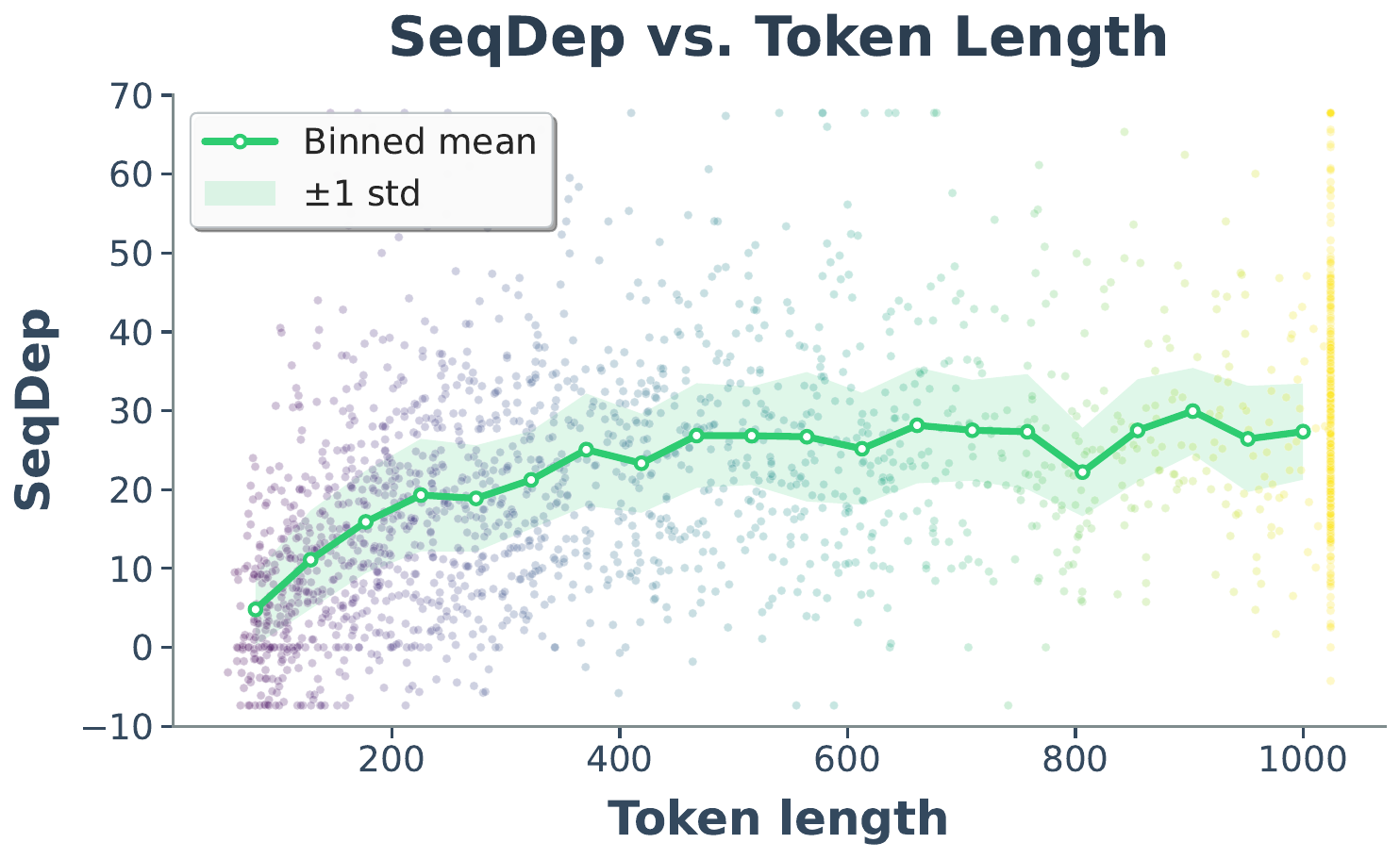}
        \caption{Fineweb SeqDep}
        \label{fig:seqdep_fineweb}
    \end{subfigure}
    \vspace{2mm}
    \caption{
        \textbf{Sequential Dependence (SeqDep) Analysis on (a) OpenR1-Math and (b) FineWeb Datasets.} 
        The consistently high and rising SeqDep scores indicate that standard training corpora possess strong intrinsic sequentiality, driving models to internalize AR-like dependencies.}
    
    \label{fig:teaser_seqdep}
\end{figure*}
To test this conjecture, we conduct a systematic analysis of the decoding behavior of commonly used DLMs. The main findings are summarized below.

\textbf{I. Widely used training corpora are strongly sequential.}
 We quantify the sequential dependency of datasets by measuring how strongly the token at one position is determined by its preceding context. We show a consistent trend: commonly used pre-training corpora (i.e., FineWeb \cite{penedo2024finewebdatasetsdecantingweb}) and long CoT reasoning datasets (i.e., Open-R1-Math \cite{openr1_math_220k_2025}) display strong sequence dependence.
  
   \textbf{II. DLM decoding remains largely autoregressive.}
   Across widely used DLM families such as LLaDA \cite{nie2025llada} and Dream \cite{ye2025dream}, ARness remains high: the model still tends to ``lock in'' decisions in a quasi-left-to-right pattern, despite the nominally arbitrary decoding rules. Conversely, forcing genuinely low ARness behavior, for instance, by randomizing the update order aggressively, can reduce ARness but typically causes reasoning performance to collapse. Taken together, these results indicate a non-trivial tradeoff: in standard setups, either ARness stays high to maintain capacity, or lowering ARness breaks reasoning.

  \textbf{III. Training on long CoT data escalates ARness.}
Continued post-training on standard long CoT datasets further increases ARness over time. While DLMs trained from scratch (e.g., LLaDA) tend to exhibit lower ARness than those adapted from pre-trained AR models (e.g., Dream), this gap gradually narrows with sustained CoT supervision. Intuitively, long CoT supervision provides an explicit step-by-step trajectory with a privileged ordering. Matching such training targets rewards the model for producing and stabilizing earlier tokens before later ones, thereby progressively shifting the learned decoding dynamics toward increasingly autoregressive behavior.


\textbf{IV. Recent parallel fast-DLM methods gain speed by \emph{amplifying}, not removing, AR-like generation.}
Despite being motivated by parallel decoding, many recent fast-DLM approaches achieve practical speedups by reinforcing an underlying autoregressive computation pattern. In particular, they rely on increasingly confident early predictions or staged block-wise updates that stabilize prefixes before allowing limited parallelism downstream. As a result, parallelism is effectively gated by an AR-like convergence order, and the achieved acceleration stems from exaggerating this sequential structure rather than eliminating it.

The above findings suggest that even though the DLMs permit arbitrary decoding strategy, as DLMs are trained on highly sequentially structured data, the model tends to internalize an AR-like computational strategy. In other words, the training distribution teaches the model that reasoning is a \emph{chain} with a privileged order, and changing the decoding procedure alone is often insufficient to undo this learned reliance. Addressing the issue therefore requires revisiting the data and supervision that shape the model’s generation strategy in the first place.

To this end, we propose \textbf{NAP} (\textit{Non-Autoregressive Parallel DLMs}), a \textbf{proof-of-concept} approach that tackles the problem from a data and decoding codesign perspective. 
First, we curate supervision in which each example consists of multiple \emph{independent reasoning trajectories} generated in parallel, this format deemphasizes any privileged token order and is naturally compatible with denoising-style learning in DLMs. Second, we introduce a \emph{parallel-forced} decoding strategy that explicitly encourages multi-token parallel updates at different reasoning traces, further steering generation away from AR-like critical paths.
Together, these two components provide a simple and effective way to better align DLM behavior with truly parallel decoding. Across a range of math reasoning benchmarks, our results show that NAP, fine-tuned with 103K samples, consistently yields stronger performance under parallel decoding than the baseline trained on standard long CoT datasets. Moreover, the improvement becomes more pronounced as we increase the degree of parallelism, indicating that NAP is better aligned with non-AR decoding dynamics rather than relying on an implicit sequential critical path. 

Note that our goal is \textbf{not} to claim that NAP fully resolves the challenges of non-AR parallel decoding. Rather, we aim to use this small-scale post-training only result to show that revisiting data and supervision design is a promising direction for mitigating AR-like behavior in DLMs and moving toward genuinely non-autoregressive parallel generation. We hope our results motivate further work on data-centric approaches to unlock the full efficiency potential of DLMs.

\section{Related Work}

\subsection{Diffusion Language Models}

Diffusion models~\citep{DBLP:journals/corr/Sohl-DicksteinW15,DBLP:conf/nips/HoJA20,DBLP:conf/iclr/0011SKKEP21}, best known for their success in image generation~\citep{DBLP:conf/cvpr/RombachBLEO22,DBLP:conf/icml/NicholDRSMMSC22,DBLP:conf/nips/SahariaCSLWDGLA22}, are increasingly studied as a non-autoregressive alternative for text generation. Bringing diffusion from continuous variables to discrete tokens can be formalized by treating the forward corruption as a Markov process over a finite vocabulary: D3PM~\citep{DBLP:conf/nips/AustinJHTB21} instantiates this idea with discrete-time transition matrices, while subsequent work extends it to continuous time through CTMC formulations~\citep{DBLP:conf/nips/CampbellBBRDD22}. A particularly practical family is masked diffusion, which can be viewed as an absorbing-state construction in the D3PM lineage and operates directly in token space via random masking~\citep{DBLP:conf/nips/ShiHWDT24}. This paradigm has produced strong results across scales, from smaller models such as MDLM~\citep{DBLP:conf/nips/SahooASGMCRK24} and RADD~\citep{DBLP:conf/iclr/OuNXZSLL25} to large systems like LLaDA~\citep{DBLP:journals/corr/abs-2502-09992} and Dream~\citep{ye2025dream}. Beyond text-only settings, MMaDA~\citep{DBLP:journals/corr/abs-2505-15809} further generalizes large diffusion models to multimodal generation with a shared probabilistic view and modality-agnostic architecture, while the broader literature highlights potential benefits such as parallelizable decoding and flexible (non left-to-right) generation orders that may be useful for complex reasoning.

\subsection{Decoding Order and Sampling Schedules}

A key degree of freedom in masked diffusion language models is the {sampling path}---which positions are updated (or committed) at each refinement step and in what order. Rather than being a mere implementation detail, several works treat order as an explicit control knob for quality/efficiency trade-offs. P2~\citep{peng2025path} cast order selection as a planning problem, where a separate planner chooses which tokens to denoise at each step, decoupling {where/when to update} from {how to update}. Prophet~\citep{li2025diffusion} further leverages model confidence to {early-commit}, switching from iterative refinement to one-shot completion when the top-2 gap indicates convergence. Order-awareness has also been pushed into training, e.g., by encouraging simpler and more coherent sampling paths~\citep{zhuspmdm}. Meanwhile, \citet{ni2026flexibility} caution that arbitrary-order flexibility can be a double-edged sword: models may preferentially resolve low-uncertainty tokens and bypass high-uncertainty branching points, collapsing the effective reasoning space, suggesting that constraining or regularizing generation order can sometimes improve reasoning.

\section{Preliminaries}

\label{sec:preliminaries}

\subsection{Diffusion Language Models}
\label{sec:prelim:dlm}

We consider {diffusion language models} (DLMs), and in particular {masked diffusion models} (MDMs), which generate discrete token sequences by iteratively denoising a partially masked state.
Let $x$ denote the input prompt and let $y_0=(y_0^1,\dots,y_0^L)\in\mathcal{V}^L$ denote a clean output sequence of length $L$ over vocabulary $\mathcal{V}$.
MDMs define a forward {masking} process indexed by a continuous time variable $t\in[0,1]$, where $t$ represents the {masking ratio}.
Given $y_0$, the forward process independently masks each token with probability $t$:
\begin{equation}
q\!\left(y_t^i \mid y_0^i\right)=
\begin{cases}
\texttt{[MASK]}, & \text{with prob. } t,\\
y_0^i, & \text{with prob. } 1-t,
\end{cases}
\label{eq:mdm-forward}
\end{equation}
and factorizes across positions as $q(y_t\mid y_0)=\prod_{i=1}^L q(y_t^i\mid y_0^i)$.
At $t=1$, the sequence is fully masked; at $t=0$, it remains unchanged.

\subsection{Measuring Autoregressive Bias}
\label{sec:prelim:arness}

To quantify how autoregressive-like a DLM decoding trajectory is, we adopt the \textbf{ARness} metrics proposed by~\citet{gong2025diffucoder}, which distinguish between global left-to-right bias and local sequential continuity. Let the decoding process be represented by a sequence of unmasked positions $\mathbf{p} = (p_1, p_2, \dots, p_L)$, where $p_c \in \{1, \dots, L\}$ denotes the position index of the token committed at decoding step $c$. Let $M_{c-1}$ be the set of masked positions just before step $c$.

\paragraph{Global ARness.}
This metric measures the tendency to prioritize unmasking the leftmost remaining tokens, capturing a front-to-back filling strategy. For a tolerance window $k \ge 1$, we define an indicator $\mathbb{I}_{\text{global}}(c, k)$ that is $1$ if the chosen position $p_c$ is among the $k$ earliest positions in $M_{c-1}$:
\begin{equation}
\mathbb{I}_{\text{global}}(c, k) =
\begin{cases}
1, & \text{if } p_c \in \text{smallest-}k(M_{c-1}), \\
0, & \text{otherwise}.
\end{cases}
\end{equation}
The Global ARness score is the average over the sequence:
\begin{equation}
\text{Global-ARness}@k = \frac{1}{L} \sum_{c=1}^L \mathbb{I}_{\text{global}}(c, k) \in [0, 1].
\label{eq:global_arness}
\end{equation}
A score of 1.0 (at $k=1$) indicates a strict autoregressive (left-to-right) generation order.

Unless otherwise stated, we use \textbf{Global-ARness@1} as the primary measure of ARness in our analysis, as it directly quantifies the adherence to a causal generation order.
\subsection{Measuring Sequential Dependence (SeqDep)}
\label{sec:prelim:seqdep} 

To quantify the intrinsic sequentiality of a dataset, we measure how much the prediction of a current text segment relies on its preceding generation history compared to relying solely on the initial prompt. Let $x$ denote the input prompt. Suppose the corresponding output sequence is divided into $N$ segments $\mathbf{s} = (s_1, \dots, s_N)$. Using an external autoregressive scorer $p_{\mathrm{AR}}$ (e.g., a pretrained LLM), we define the Sequential Dependence (SeqDep) as the average log-probability gain provided by the prefix context:
\begin{equation}
\begin{split}
    \mathrm{SeqDep}(x, \mathbf{s}) &= \frac{1}{N-1}\sum_{n=2}^N \Big( \log p_{\mathrm{AR}}(s_n \mid x, s_{<n}) \\
    &\quad - \log p_{\mathrm{AR}}(s_n \mid x) \Big)
\end{split}
\label{eq:seqdep}
\end{equation}

Intuitively, this metric captures the conditional dependence between reasoning steps. A $\mathrm{SeqDep}$ score near $0$ indicates that the segment $s_n$ is conditionally independent of previous segments $s_{<n}$ given the prompt $x$, implying that the sequence components could theoretically be generated in parallel. Conversely, a high positive $\mathrm{SeqDep}$ score indicates a strong chain-like structure—meaning later tokens are heavily dictated by the immediate preceding context, a hallmark of standard left-to-right autoregressive reasoning.

\section{Decoding Behaviors of DLMs}
\label{sec:motivation}

In this section, we conduct a systematic analysis of the decoding behavior of commonly used DLMs. We fix the pretrained masked diffusion model, the token budget, the number of refinement steps, and the
mask-ratio schedule, and vary {only} the decoding rule.
This isolates the effect of the induced generation order from all other factors. Our main findings are summarized below.

\subsection{Strong Sequential Dependence in Training Corpora}
\label{sec:motivation:data}
A primary driver of sequential behavior is the data itself. We hypothesize that if the training distribution is highly sequential, the model learns an implicit left-to-right dependency that persists even under parallel decoding objectives.

We quantify this using the $\mathrm{SeqDep}$ metric (Sec.~\ref{sec:prelim:seqdep}) on two representative datasets: {FineWeb} (pre-training corpora) and {OpenR1-Math} (long-CoT reasoning).
As shown in Figure~\ref{fig:seqdep_fineweb} and ~\ref{fig:seqdep_openmath}, both datasets display strong sequence dependence. Notably, reasoning steps in OpenR1-Math exhibit increasing dependence as the chain progresses ($p_{\mathrm{AR}}$ predicts later steps with much higher confidence given the prefix).
This suggests that standard training data teaches the model that reasoning is a fundamentally ordered chain, creating a mismatch with position-agnostic diffusion objectives.

\subsection{DLMs' Decoding Remains Largely Autoregressive}
\label{sec:motivation:behavior}

\looseness=-1 Given sequential training data, we examine how DLMs behave during inference. We evaluate two popular models, LLaDA-8B~\cite{nie2025llada} and Dream-7B~\cite{ye2025dream}, under three distinct decoding strategies:
(i) \textbf{Autoregressive (AR) Order}: committing the leftmost unresolved tokens at each step, mimicking standard left-to-right generation; 
(ii) \textbf{Arbitrary Order (AO)}: a confidence-based strategy that commits the most certain tokens first regardless of their positions; 
and (iii) \textbf{Random}: committing a uniformly random subset of tokens at each step.

Unlike Dream-7B, which handles fully unstructured parallel updates relatively well, LLaDA-8B exhibits severe degradation on structured mathematical tasks if unmasked entirely at random. This is largely an artifact of its specific supervised fine-tuning (SFT) phase. To ensure valid and comparable generation quality for LLaDA, we apply a constrained \textit{block-wise} modification~\cite{arriola2025block} to the AO and Random strategies.

\begin{table}[h]
\centering
\caption{
\textbf{Quantifying Autoregressive Bias (ARness) and Accuracy.} 
Comparison of sequential bias and performance across different decoding strategies.
While AR Order implies strict sequentiality (1.00), AO (Conf) maintains high ARness and competitive accuracy.
}
\label{tab:motivation:arness_acc}
\resizebox{\linewidth}{!}{
\begin{tabular}{l cc c cc}
\toprule
\multirow{2}{*}{\textbf{Model}} & \multicolumn{2}{c}{\textbf{AR Order}} & & \multicolumn{2}{c}{\textbf{AO (Conf)}} \\
\cmidrule{2-3} \cmidrule{5-6}
 & \textbf{ARness} & \textbf{Acc} & & \textbf{ARness} & \textbf{Acc} \\
\midrule
LLaDA-8B~\cite{nie2025llada}       & 1.00 & 71.9 & & 0.73 & 51.9 \\
+ Fast-dLLM~\cite{wu2025fast}& 1.00 & 71.9 & & \bf 0.87 & 51.6 \\
\midrule
Dream-7B~\cite{ye2025dream}        & 1.00 & 78.2 & & 0.92 & 78.2 \\
+ Fast-dLLM~\cite{wu2025fast}& 1.00 & 78.3 & & \bf 0.94 & 78.1 \\
\bottomrule
\end{tabular}
}
\end{table}

\textbf{High ARness in DLM Decoding.}
Table~\ref{tab:motivation:arness_acc} reports the ARness scores. While AR order is 1.0 by definition, {AO decoding converges to extremely high ARness} ($\sim0.92$ for Dream), indicating that the model's most ``confident" tokens are almost always the next tokens in the sequence. As a result, DLMs exhibit behavior closely resembling autoregressive generation.

\textbf{The Accuracy--ARness Tradeoff.}
\looseness=-1 Is it possible to force genuinely parallel behavior? We test this using a Random decoding strategy, which successfully yields near-zero ARness. However, as shown in Figure~\ref{fig:teaser_accuracy}, this comes at a severe cost: reasoning accuracy on \textsc{GSM8K}~\cite{cobbe2021gsm8k} and \textsc{MATH 500}~\cite{lightman2023lets} collapses when the model is prevented from following a sequential path. These results suggest that strong reasoning performance is often obtained at the cost of genuine parallelism, as improved accuracy tends to coincide with higher \emph{AR-ness} under standard setups.

\subsection{Long-CoT Supervision Escalates AR-ness}
\label{sec:motivation:cot}

\begin{table}[t]
    \centering
    \caption{\textbf{Long-CoT Supervision Increases ARness.} Comparison of Global ARness@1 scores (using AO decoding) before and after fine-tuning.}
    \label{tab:cot_arness}
    \small
    \resizebox{0.99\linewidth}{!}{
    \begin{tabular}{l c c c}
        \toprule
        \textbf{Model} & \textbf{Base (Pretrained)} & \textbf{Long-CoT (SFT)} & \textbf{Change} \\
        \midrule
        LLaDA-8B & 0.73 & 0.81 & \textcolor{red}{$\uparrow$ 0.08} \\
        Dream-7B & 0.92 & 0.93 & \textcolor{red}{$\uparrow$ 0.01} \\
        \bottomrule
    \end{tabular}}
\end{table}

\begin{figure}[t]
    \centering
    \includegraphics[width=0.9999\linewidth]{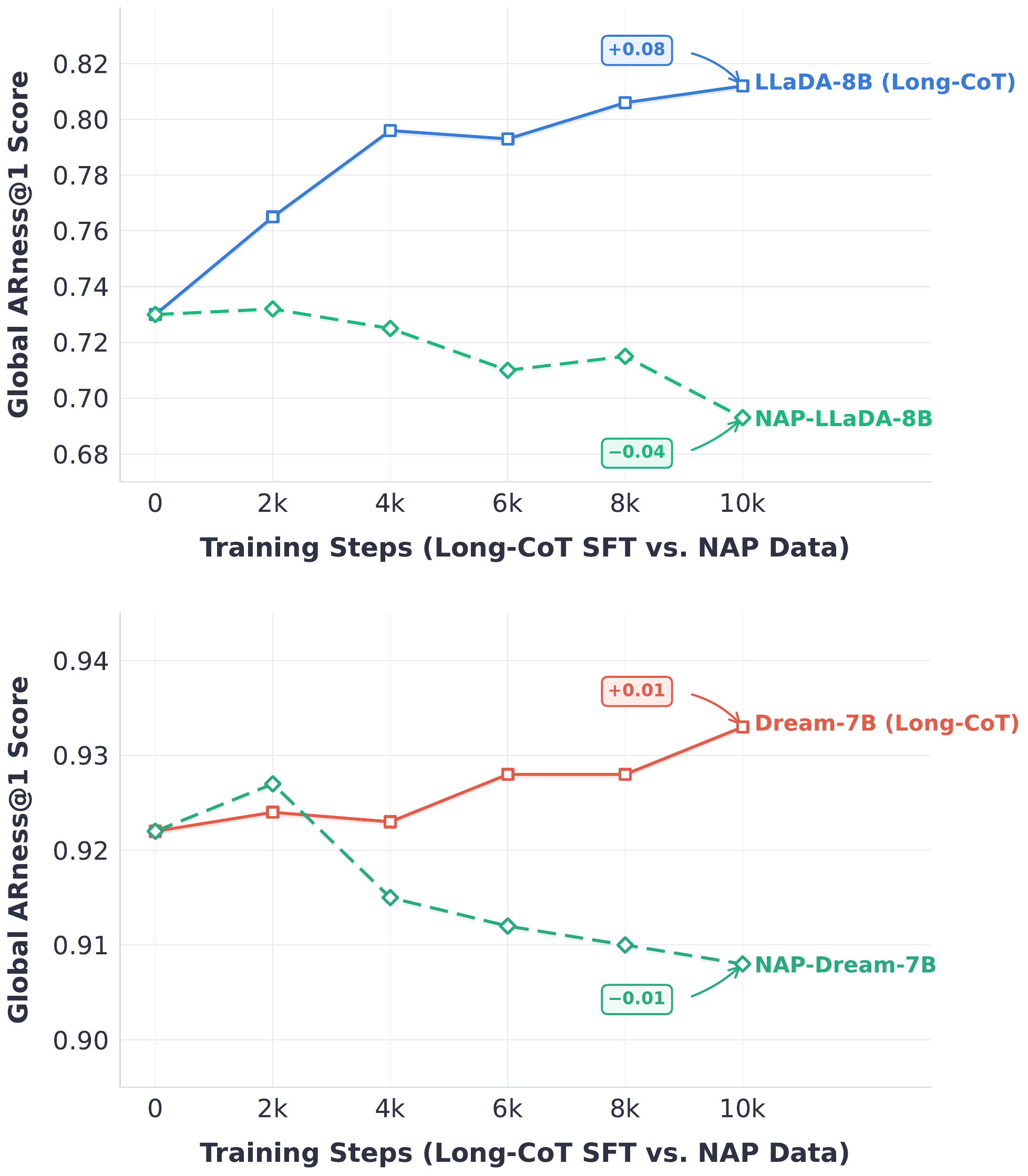}
    \caption{\textbf{Long-CoT Supervision Increases ARness.} 
    The positive deltas show models converging toward strict left-to-right generation (1.0), confirming that current supervision methods actively discourage non-autoregressive parallel decoding.}
    \label{fig:cot_dynamics}
\end{figure}

We further investigate how supervised fine-tuning (SFT) on long Chain-of-Thought (CoT)  data affects decoding dynamics.
We compare the ARness of base models against checkpoints post-trained on standard CoT datasets (Open-R1 Math \citep{openr1_math_220k_2025}).

As shown in Table~\ref{tab:cot_arness} and Figure~\ref{fig:cot_dynamics}, results indicate a clear trend: \textbf{post-training further increases ARness}. 
For instance, LLaDA's base ARness under AO decoding rises from 0.73 to 0.81 after CoT tuning. 
Intuitively, CoT supervision provides explicit step-by-step trajectories with a privileged order. Minimizing the loss on such data rewards the model for stabilizing earlier tokens before later ones, effectively "baking in" the AR order and making it harder for the model to utilize genuine parallel decoding during inference.

\subsection{Current Fast DLMs Reinforce Sequentiality}
\label{sec:motivation:fast}
Finally, we analyze whether specialized "fast" decoding algorithms can unlock genuine parallelism. We evaluate \textbf{Fast-dLLM}~\cite{wu2025fast}, a state-of-the-art acceleration method that employs block-wise parallel decoding.

As shown in Table~\ref{tab:motivation:arness_acc}, these methods do not reduce sequential dependence; in fact, they exacerbate it. For instance, while standard AO decoding for LLaDA has an ARness of 0.73, applying Fast-dLLM pushes this score up to \textbf{0.87}. Similarly, for Dream-7B, the ARness rises to \textbf{0.94}, nearly indistinguishable from strict autoregressive decoding (1.00).
\begin{figure}[h]
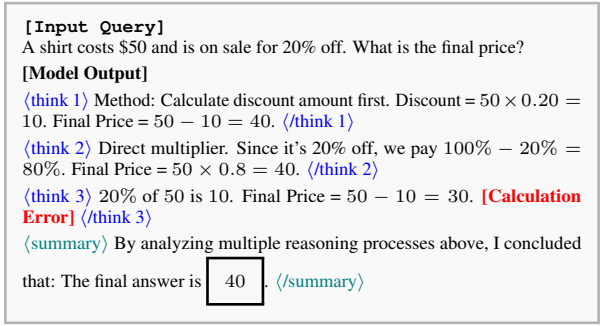

    \centering
    \begin{small}
    \setlength{\fboxsep}{6pt}
    \setlength{\fboxrule}{1pt}
    \fcolorbox{black!30}{gray!5}{%
    \begin{minipage}{0.9\linewidth}
    \ttfamily
    \scriptsize
    
    \textbf{[Input Query]}
    
    \normalfont
    A shirt costs \$50 and is on sale for 20\% off. What is the final price?
    
    \vspace{0.3em}
    \textbf{[Model Output]}
    
    \vspace{0.3em}
    \textcolor{blue}{$\langle$think 1$\rangle$}
    \normalfont
    Method: Calculate discount amount first.
    Discount = $50 \times 0.20 = 10$.
    Final Price = $50 - 10 = 40$.
    \textcolor{blue}{$\langle$/think 1$\rangle$}
    
    \vspace{0.3em}
    \textcolor{blue}{$\langle$think 2$\rangle$}
    \normalfont
    Direct multiplier.
    Since it's 20\% off, we pay $100\% - 20\% = 80\%$.
    Final Price = $50 \times 0.8 = 40$.
    \textcolor{blue}{$\langle$/think 2$\rangle$}
    
    \vspace{0.3em}
    \textcolor{blue}{$\langle$think 3$\rangle$}
    \normalfont
    $20\%$ of $50$ is $10$.
    Final Price = $50 - 10 = 30$. \textcolor{red}{\textbf{[Calculation Error]}}
    \textcolor{blue}{$\langle$/think 3$\rangle$}
    
    \vspace{0.3em}
    \textcolor{teal}{$\langle$summary$\rangle$}
    \normalfont
    By analyzing multiple reasoning processes above, I concluded that: The final answer is \boxed{40}.
    \textcolor{teal}{$\langle$/summary$\rangle$}
    
    \end{minipage}%
    }
    \end{small}
    \vspace{-5pt}
    \caption{A compact training instance. The model generates parallel paths (including distinct methods and a noisy path) and aggregates them into a correct summary.}
    \label{fig:short_example}
\end{figure}
\looseness=-1 This empirical evidence suggests that current "fast" DLMs achieve speedups not by enabling non-sequential generation, but by effectively identifying and accelerating the underlying autoregressive critical path. The parallelism in these systems is gated by the convergence of the prefix, meaning they optimize the execution of the sequential chain rather than eliminating the bottleneck. This diagnosis reinforces our core premise: achieving \textit{true} non-autoregressive parallelism requires revisiting the supervision signal itself, rather than relying solely on inference-time algorithmic optimizations.

\section{NAP: Non-Autoregressive Parallel DLMs}
\label{sec:method}

\begin{figure*}[t]
    \centering
    \includegraphics[width=1.0\textwidth]{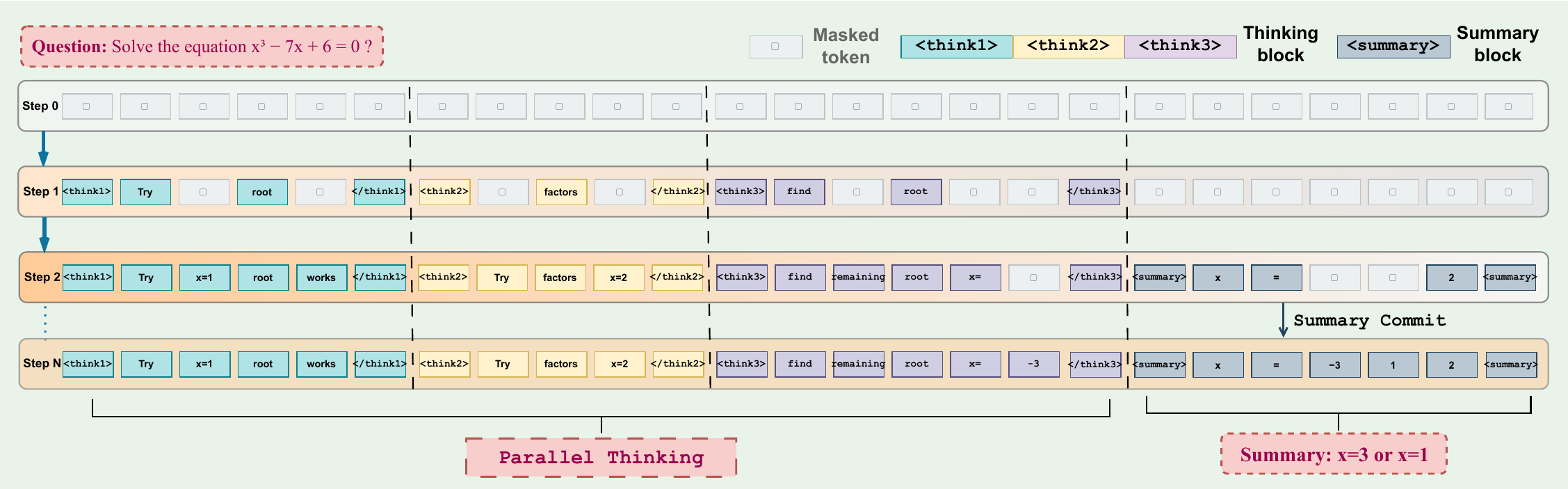}
    \caption{\textbf{Overview of the parallel-forced decoding framework}. The model concurrently generates multiple independent reasoning paths within structured thinking blocks. These parallel trajectories are then synthesized into a result within a designated summary block.}
    \label{fig:pipeline}
\end{figure*}

\subsection{Overview}
\label{sec:method:overview}

To bridge the gap between DLM objectives and the sequential nature of reasoning data, we propose \textbf{NAP} (\textit{Non-Autoregressive Parallel DLMs}). NAP is a data-decoding co-design framework that breaks the implicit autoregressive lock-in by restructuring both the supervision signal and the inference process. The framework operates on two levels: first, it curates training examples as multiple {independent reasoning trajectories} rather than a single linear chain, thereby removing the notion of a privileged order; second, it employs a \textbf{parallel-forced decoding} strategy that explicitly enforces multi-stream updates during inference, preventing the model from collapsing into a sequential critical path.

\subsection{Data Curation}
\label{sec:method:data}
Standard chain-of-thought (CoT) data typically encodes a single canonical left-to-right reasoning order, creating a natural mismatch with the objective of parallel DLM decoding. To address this, we curate a dataset $\mathcal{D}_{\text{parallel}}$ whose supervision is {inherently parallel}.
\vspace{-2mm}
\paragraph{Generating Parallel Reasoning Traces.}
Similar to ParaThinker~\cite{wen2025parathinker}, given a query $x$, we prompt a strong teacher model to generate $P$ independent reasoning traces $\{r^{(1)}, \dots, r^{(P)}\}$. We employ a high sampling temperature ($\tau=1.0$) to induce diverse problem-solving approaches or distinct logical orderings. Unlike standard augmentation which treats these as separate samples, NAP groups them into a single training instance. This ensures that the parallel paths represent truly independent explorations rather than redundant copies.

\paragraph{Summary and Aggregation.}
To teach the model to resolve conflicts, we construct the summary block $S$ by conditioning the ground-truth answer $a$ on the concatenation of these diverse (and potentially noisy) paths. The final training instance follows the format in Eq.~\eqref{eq:pas-template}, {as illustrated in Figure~\ref{fig:short_example}}.
In this setup, the model observes multiple parallel paths—some of which may contain errors (e.g., Path 3)—followed invariably by the correct result in $S$. This supervision forces the model to implicitly learn how to identify valid reasoning streams and filter out noise to match the ground truth, treating the parallel paths as supporting evidence rather than a linear chain. We fine-tune the DLM on this structured data using the standard masked diffusion objective.
\subsection{Parallel-Forced Decoding}
\label{sec:method:decoding}

To enable the model to reason in parallel, we design a decoding canvas that spatially separates reasoning streams and enforce a structure-aware update schedule.

\paragraph{Decoding Canvas.}
We define a structured output format containing $m$ independent reasoning blocks and one summary block:
\begin{equation}
    Y \;=\; \big[\, B_1, R^{(1)},\; B_2, R^{(2)},\; \dots,\; B_m, R^{(m)},\; B_\mathrm{S}, S \,\big],
    \label{eq:pas-template}
\end{equation}
where $B_j$ are fixed textual headers (e.g., ``\texttt{<think \#j>}''), $R^{(j)}$ are free-form reasoning contents for the $j$-th path, and $S$ is a final summary containing the answer. Given a prompt $x$, we initialize a canvas of length $L = \sum (|B_j| + L_j) + (|B_\mathrm{S}| + L_\mathrm{S})$, where fixed headers are clamped and reasoning slots are initialized to \texttt{[MASK]}. This layout effectively enforces \textit{conditional independence} between $R^{(i)}$ and $R^{(j)}$ given the prompt, as there is no causal masking order between them in a bidirectional model.

\paragraph{Macro-Parallel, Micro-Confidence Updates.}
\looseness=-1 Standard arbitrary-order decoding often degenerates into global sequential generation because the model preferentially resolves the immediate next tokens. NAP-D prevents this via a hierarchical schedule. At the {macro level}, we enforce strict parallelism: the unmasking budget is distributed across \textit{all} $m$ reasoning blocks $\{R^{(1)}, \dots, R^{(m)}\}$ at every step. This constraint prevents the model from stabilizing upstream paths before initiating downstream ones. At the {micro level}, within each individual block $R^{(j)}$, we apply a {confidence-based strategy} (i.e., masking low-confidence tokens). We do not enforce a left-to-right order locally; instead, tokens are committed based on their confidence scores. This combination ensures that the global process is parallel (evolving multiple trajectories simultaneously) while local generation retains the flexibility of non-autoregressive refinement.

\section{Experiments}

\begin{table*}[t]
  \centering
  \caption{Benchmark results on LLaDA-8B-Instruct and Dream-7B-Instruct under different step budgets. Tok/Step denotes the number of tokens decoded per decoding step; larger Tok/Step corresponds to higher decoding parallelism.}
  \label{tab:main_results}
  \renewcommand{\arraystretch}{1.22}
  \resizebox{0.99\textwidth}{!}{%
  \begin{tabular}{l c c |c c O|c c O}
    \toprule
    \textbf{Benchmark} & \textbf{Steps} & \textbf{Tok/Step} &
    \texttt{LLaDA 8B} &
    \texttt{LLaDA 8B (Long-CoT)} &
    \textbf{NAP-LLaDA 8B} &
    \texttt{Dream-7B} &
    \texttt{Dream-7B (Long-CoT)} &
    \textbf{NAP-Dream-7B} \\
    \midrule

    \multicolumn{9}{c}{\textbf{Mathematics \& Scientific}} \\
    \midrule

    \multirow{4}{*}{GSM8K}
      & 256  & 4 & 46.4 & 54.1 & \textbf{56.1}~\improve{+2.0} & 35.0 & 46.5 & \textbf{60.9}~\improve{+14.4} \\
      & 336  & 3 & 54.4 & 60.9 & \textbf{63.3}~\improve{+2.4} & 49.4 & 56.9 & \textbf{70.9}~\improve{+14.0} \\
      & 512  & 2 & 62.0 & 82.0 & \textbf{82.6}~\improve{+0.4} & 58.5 & 66.8 & \textbf{79.2}~\improve{+12.4} \\
      & 1024 & 1 & 66.5 & 83.5 & \textbf{84.1}~\improve{+0.6} & 68.9 & 78.0 & \textbf{83.6}~\improve{+5.6} \\
    \midrule

    \multirow{4}{*}{MATH-500}
      & 256  & 4 & 17.8 & 21.4 & \textbf{26.6}~\improve{+5.2} & 8.8 & 16.2 & \textbf{23.8}~\improve{+7.6} \\
      & 336  & 3 & 20.6 & 26.6 & \textbf{35.4}~\improve{+8.8} & 11.4 & 25.6 & \textbf{31.4}~\improve{+5.8} \\
      & 512  & 2 & 28.0 & 41.2 & \textbf{43.0}~\improve{+1.8} & 20.8 & 40.0 & \textbf{43.0}~\improve{+3.0}\\
      & 1024 & 1 & 30.4 & 45.0 & \textbf{47.0}~\improve{+2.0} & 35.0 & 47.4 & \textbf{49.6}~\improve{+2.2} \\
    \midrule

    \multirow{4}{*}{GPQA}
      & 336  & 3 & 12.5 & 15.4 & \textbf{19.0}~\improve{+3.6} & 5.8 & 7.3 & \textbf{10.5}~\improve{+3.2} \\
      & 512  & 2 & 18.8 & 21.2 & \textbf{25.9}~\improve{+4.7} & 14.7 & 19.4 & \textbf{22.5}~\improve{+3.1} \\
      & 1024 & 1 & 20.8 & 23.0 & \textbf{28.6}~\improve{+5.6} & 26.1 & 28.6 & \textbf{29.5}~\improve{+0.9} \\

    \bottomrule
  \end{tabular}%
  }
\end{table*}

\label{sec:experiments}
This section evaluates whether our decoding strategy can (i) improve reasoning performance over standard diffusion decoding rules, (ii) reshape the induced generation order as measured by ARness (Section~\ref{sec:prelim:arness}), and (iii) mitigate order sensitivity in regimes where long-form rationales exhibit strong sequential dependence (Eq.~\eqref{eq:seqdep}). Unless otherwise stated, \emph{all} results use the same pretrained masked diffusion model and differ \emph{only} in the decoding rule.

\textbf{Evaluation protocol.}
We evaluate on a suite of reasoning benchmarks including \textsc{GSM8K}~\cite{cobbe2021gsm8k}, \textsc{MATH-500}~\cite{lightman2023lets}, and \textsc{GPQA}~\cite{rein2024gpqa}.
Each example is prompted to produce a thinking path and a final answer in a fixed format; we extract answers with a deterministic parser and report accuracy.

\textbf{Models and Training.}
We conduct experiments on two state-of-the-art diffusion language models: \texttt{LLaDA-8B-Instruct}~\cite{nie2025llada} and \texttt{Dream-7B-Instruct}~\cite{ye2025dream}.
To validate our proposed method, we fine-tune these base models on the parallel reasoning dataset $\mathcal{D}_{\text{parallel}}$ curated via the pipeline described in Section~\ref{sec:method:data}.
For a fair comparison, we also train a {Long-CoT} baseline on the same set of reasoning trajectories but serialized in the standard autoregressive format. Crucially, this baseline is evaluated using standard decoding—its optimal inference setting—rather than our parallel strategy, ensuring a strong and fair comparison.
Both variants are trained using the standard masked diffusion objective for 3 epochs.
We use the AdamW optimizer with a learning rate of 2e-6 and a global batch size of 256.
All experiments are conducted on 8 NVIDIA A800 GPUs.

\textbf{Decoding baselines.}
We compare several widely used unmasking rules under the common mask-and-predict framework.
\textbf{AR order} commits the leftmost unresolved tokens at each step (a diffusion realization of left-to-right decoding).
\textbf{Arbitrary order (AO)} commits the most confident positions.
\textbf{Random order (Rand)} commits a uniformly random subset at each step, serving as a low-ARness control.
Our method generates $m$ multiple independent reasoning paths and a final summary commit on a structured canvas.
To ensure a fair budget, PaS-Dec uses the same total token cap $L$ by allocating per-path budgets $330$ and a summary budget $32$ such that the overall canvas length matches the baseline.
The summary block is the only region used for answer extraction and scoring.

\subsection{Main Results}
Table~\ref{tab:main_results} summarizes the performance across three benchmarks. Across all benchmarks and step budgets, our method achieves higher accuracy than both the Base model and the Long-CoT baseline. For instance, on GSM8K with Dream-7B (1024 steps), NAP-Dream-7B reaches {83.6\%}, surpassing the Long-CoT model (78.0\%) despite using the same amount of compute and training data. This suggests that organizing reasoning into parallel streams is a more effective supervision signal for DLMs than forcing a single long chain.

The most significant advantage of NAP appears in the low-step regime, e.g., 256 steps (4x parallel), where the model must generate more than one token per forward pass. Standard \texttt{Long-CoT} models degrade sharply as parallelism increases. On Dream-7B/GSM8K, accuracy drops from 78.0\% (1024 steps) to 46.5\% (256 steps). This confirms that standard supervision creates a dependency on sequential stability; when forced to hurry, the reasoning collapses. In the same setting, NAP-Dream-7B maintains strong accuracy at 60.9\%, compared to 46.5\% of the Long-CoT baseline, thereby retaining substantially more capability. Notably, the gap between NAP and Long-CoT widens as parallel decoding is made more aggressive, increasing from +5.6\% at 1024 steps to \textbf{+14.4\%} at 256 steps.
This result validates our core hypothesis: by training on data that lacks a privileged order, the model learns to be less reliant on the immediate left-side context, enabling effective Non-AR parallel decoding.

To further understand how NAP achieves these results, we analyze the relationship between performance and the sequential nature of generation (ARness). As shown in Figure~\ref{fig:decoding_comparison}, standard models (LLaDA/Dream) using Arbitrary Order (AO) decoding exhibit a strict diagonal pattern. Even though they can decode anywhere, they effectively collapse into a left-to-right process (High ARness). In contrast, NAP (Figure~\ref{fig:decoding_comparison}(d)) displays distinct parallel bands, confirming that multiple reasoning trajectories are being generated simultaneously.

\subsection{Ablation Studies}
\label{sec:exp:ablation}

We investigate the individual contributions of the supervision data and the decoding strategy using Dream-7B on the GSM8K benchmark.

\textbf{The Necessity of Data-Decoding Co-design.}
We first isolate the impact of our proposed decoding method versus the parallel-aligned data. As shown in Table~\ref{tab:llada_math500_need_sft}, applying our Parallel-Forced Decoding strategy to a standard base model that has not been trained with our data leads to a larger performance drop than standard Arbitrary Order (AO) decoding. This suggests that without training support, the original Dream-7B struggles to handle the fragmented context of simultaneous generation. In addition, the decoding strategy becomes critical when parallelism is high. Specifically at the aggressive 256-step budget, our Parallel-Forced decoding outperforms AO (60.9\% vs. 57.4\%). This confirms that while the data provides the foundational reasoning capability, aligning the decoding strategy is essential to maintain robustness when forcing the model to generate multiple tokens in parallel.

\begin{table}[h]
  \centering
  \caption{GSM8K accuracy using Dream-7B. Simply applying parallel decoding to a base model hurts performance; gains require aligned supervision.}
  \label{tab:llada_math500_need_sft}
  \small
  \begin{tabular}{l l c c c}
    \toprule
    \textbf{Training Data} & \textbf{Decoding} &
    \textbf{256} &
    \textbf{512} &
    \textbf{1024} \\
    \midrule
    Base (Pretrained) & AO & 35.0 & 58.5 & 68.9 \\
    Base (Pretrained) & \textbf{Parallel-Forced} & 31.0 & 52.6 & 60.2 \\
    \midrule
    \textbf{NAP (Ours)} & AO & 57.4 & 78.9 & \textbf{85.1} \\
    \textbf{NAP (Ours)} & \textbf{Parallel-Forced} & \textbf{60.9} & \textbf{79.2} & {83.6} \\
    \bottomrule
  \end{tabular}
\end{table}

\textbf{Impact of Parallel Width ($m$).}
We further analyze how the number of parallel reasoning paths affects performance while keeping the total token budget constant. As detailed in Table~\ref{tab:ablation_paths_inline}, increasing the number of reasoning paths from a single chain ($m=1$) to three ($m=3$) provides consistent accuracy gains across both model families. Specifically, NAP-Dream sees a substantial improvement from 75.4\% to 83.6\%, while NAP-LLaDA rises from 79.4\% to 84.1\%. This monotonic trend supports the view that NAP benefits from an ``internal ensemble'' effect, where the final summary block effectively aggregates insights from multiple diverse trajectories generated in parallel to derive a more robust answer.

\begin{table}[h]
  \centering
  \caption{Accuracy on GSM8K with varying $m$. Total token budget is fixed.}
  \label{tab:ablation_paths_inline}
  \small
  \begin{tabular}{l c c c}
    \toprule
    \textbf{Method} & \textbf{1 Path} & \textbf{2 Paths} & \textbf{3 Paths} \\
    \midrule
{NAP-Dream} & {75.4} & {78.9} & {83.6} \\
{NAP-LLaDA} & {79.4} & {82.6} & {84.1} \\
    \bottomrule
  \end{tabular}
\end{table}

\begin{figure}[t]
    \centering
    \includegraphics[width=0.9\linewidth]{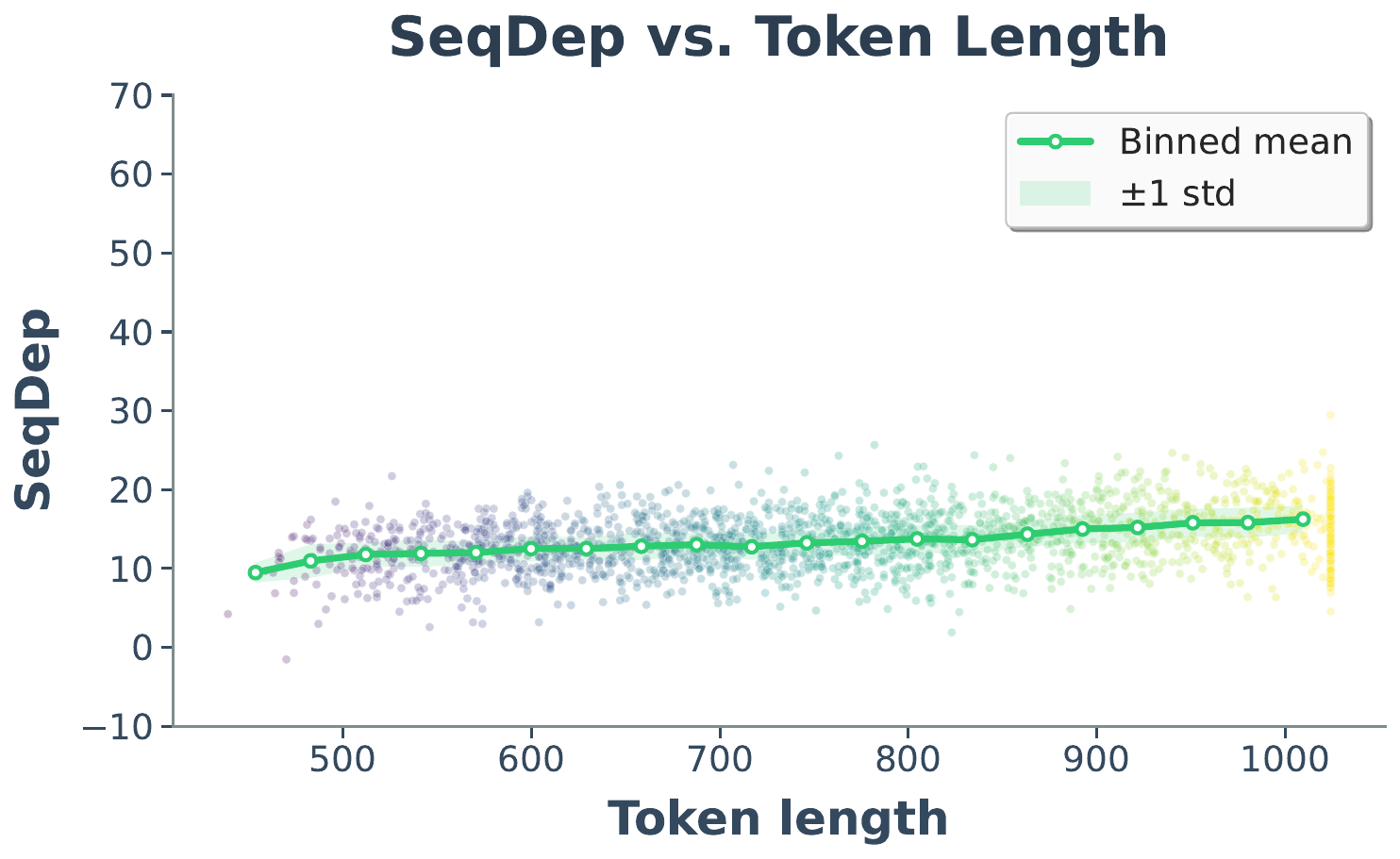}
    \caption{
    \textbf{SeqDep Analysis on $\mathcal{D}_{\text{parallel}}$.} 
    We visualize the Sequential Dependence (SeqDep) of our curated parallel reasoning data against token length. The green curve (binned mean) shows that SeqDep remains stable and relatively low across varying lengths.
    }
    \label{fig:ours_seqdep}
\end{figure}

\textbf{Intrinsic Parallelism of Curated Data.}
To verify that our data curation pipeline effectively reduces the autoregressive bottleneck, we analyze the Sequential Dependence (SeqDep) of our constructed dataset $\mathcal{D}_{\text{parallel}}$. As illustrated in Figure~\ref{fig:ours_seqdep}, the SeqDep score remains remarkably stable (mean $\approx 12$) even as the sequence length grows from 500 to over 1000 tokens.
Unlike standard long-chain reasoning (as shown in Section~\ref{sec:motivation}), where dependence often escalates with depth, our parallel-structured data maintains a consistent level of information density. This "flat" dependency profile confirms that the reasoning trajectories within our data possess high conditional independence, providing the necessary learning signal for the model to perform effective parallel updates during inference.

\section{Conclusion}

In this work, we argue that the struggle of Diffusion Language Models (DLMs) to achieve genuine parallel decoding stems largely from the implicit sequentiality of standard training data. Our proposed method, NAP, demonstrates that aligning supervision with parallel decoding dynamics effectively mitigates this autoregressive collapse. By training on parallel reasoning trajectories and enforcing multi-stream updates, NAP decouples reasoning capability from sequential order, achieving superior performance in high-parallelism regimes while significantly reducing global ARness. These results suggest that unlocking the full potential of non-autoregressive generation requires moving beyond decoding heuristics to fundamentally rethink how we structure supervision for parallel reasoning.
\paragraph{Limitations.} 
While NAP demonstrates the feasibility of aligning supervision with genuinely parallel decoding, our current implementation serves primarily as a proof-of-concept. The method is evaluated in a post-training setting on a relatively small scale ($\sim$100K samples). As scaling laws dictate much of DLMs' behavior, a broader pre-training phase with inherently non-sequential, parallel-structured data may be required to completely eliminate the autoregressive bottleneck. 
\bibliography{example_paper}
\bibliographystyle{icml2026}

\newpage
\appendix
\onecolumn

\end{document}